\documentclass[conference]{IEEEtran}
\usepackage{cite}
\usepackage{amsmath,amssymb,amsfonts}
\usepackage{algorithmic}
\usepackage{graphicx}
\usepackage{textcomp}
\usepackage{xcolor}
\usepackage{subfigure}
\usepackage{comment}
\usepackage{float}
\usepackage[utf8]{inputenc}
\usepackage{multirow}
\definecolor{DarkGreen}{rgb}{0.0, 0.5, 0.0}
\usepackage{hyperref}
\hypersetup{
    colorlinks=true,
    linkcolor=red,
    citecolor=DarkGreen,
}
    
\begin{document}

\title{Improvement in Facial Emotion Recognition using Synthetic Data Generated by Diffusion Model}

\author{
Arnab Kumar Roy$^1$, Hemant Kumar Kathania$^2$, and Adhitiya Sharma$^2$ \\
\textit{$^1$Dept. of Computer Science and Engineering, Sikkim Manipal Institute of Technology, Sikkim, India} \\
\textit{$^2$Dept. of Electronics and Communication Engineering, National Institute of Technology Sikkim, India} \\
\thanks{The authors would like to thank iHub DivyaSampark IIT Roorkee for supporting this work.}
}

\maketitle

\begin{abstract}

Facial Emotion Recognition (FER) plays a crucial role in computer vision, with significant applications in human-computer interaction, affective computing, and areas such as mental health monitoring and personalized learning environments. However, a major challenge in FER task is the class imbalance commonly found in available datasets, which can hinder both model performance and generalization. In this paper, we tackle the issue of data imbalance by incorporating synthetic data augmentation and leveraging the ResEmoteNet model to enhance the overall performance on facial emotion recognition task. We employed Stable Diffusion 2 and Stable Diffusion 3 Medium models to generate synthetic facial emotion data, augmenting the training sets of the FER2013 and RAF-DB benchmark datasets. Training ResEmoteNet with these augmented datasets resulted in substantial performance improvements, achieving accuracies of 96.47\% on FER2013 and 99.23\% on RAF-DB. These findings shows an absolute improvement of 16.68\% in FER2013, 4.47\% in RAF-DB and highlight the efficacy of synthetic data augmentation in strengthening FER models and underscore the potential of advanced generative models in FER research and applications. The source code for ResEmoteNet is available at \url{https://github.com/ArnabKumarRoy02/ResEmoteNet}
\end{abstract}

\begin{IEEEkeywords}
Facial Emotion Recognition, Augmentation, Stable Diffusion, CNN, SENets, ResNets
\end{IEEEkeywords}

\section{Introduction}
\label{sec:intro}
Facial Emotion Recognition (FER) is a specialized image recognition task that focuses on identifying emotions from facial images or videos, which presents significant challenges due to the subtlety of facial expressions and the need for precise data collection and annotation. Despite these difficulties, FER plays a critical role in applications such as mental health monitoring, where it can detect conditions like depression and anxiety, as well as in human-computer interaction and educational systems.

Recent advancements in FER have been driven by deep learning, particularly Convolutional Neural Networks (CNNs). For instance, in \cite{residualmasking} an ensemble of networks with residual masking was introduced to improve FER accuracy, while in \cite{zhang2023dual} MobileFaceNet was utilized \cite{chen2018mobilefacenets} with a Dual Direction Attention Network (DDAN) to enhance feature extraction through attention maps. A dual-stream feature extraction method was proposed, combining an image backbone and facial landmark detector with a windowed cross-attention mechanism for improved computational efficiency \cite{poster++}. Similarly, Vision Transformers (ViTs) with Multi-View Complementary Prompters (MCPs) were utilized to integrate static and landmark-aware features, further enhancing system performance \cite{s2d}.

Accurate emotion recognition in neural networks relies on well-annotated datasets, yet many freely available FER datasets face challenges such as class imbalance and small sample sizes, with classes like Happy and Neutral often being overrepresented. This issue is prevalent across various image recognition datasets and negatively impacts model performance. To mitigate this, synthetic data augmentation techniques have been introduced, with early popularity in GAN-based methods. GANs were used to enhance pneumonia classification by augmenting chest X-ray datasets with synthetic images, achieving a validation accuracy of 94.5\% \cite{srivastav2021improved}. A modified lightweight GAN model was similarly employed to generate high-quality synthetic images, effectively addressing class imbalance and leading to a classification accuracy of 99.06\% for plastic bottle detection \cite{chatterjee2022enhancement}.

Recently, diffusion models have become increasingly important for synthetic data generation. Stable Diffusion was utilized to create a skin disease dataset through controlled text prompts \cite{akrout2023diffusion}. Similarly, GLIDE, an open-source text-to-image diffusion model, was used to generate synthetic data, significantly improving classification across 17 diverse datasets \cite{he2022synthetic}. This approach demonstrated that combining synthetic data with minimal real data could achieve state-of-the-art performance and effectively pre-train large models. In FER, an augmentation method combining deep learning and genetic algorithms was introduced, focusing on enhancing the feature set from key-frame extractions rather than augmenting visual data, leading to improved performance \cite{nida2024spatial}.

In this paper, we utilize ResEmoteNet \cite{roy2024resemotenet}, a neural network architecture designed for facial emotion recognition (FER), incorporating Residual Connections and Squeeze-and-Excitation mechanisms to enhance feature extraction. Residual Connections help mitigate vanishing gradient issues, improving the training of deep models, while the Squeeze-and-Excitation mechanism refines feature maps by recalibrating channel-wise responses, allowing the network to focus on the most relevant facial image aspects. A significant challenge in FER is the class imbalance in datasets, where emotions like Happy and Neutral are overrepresented compared to Sad or Angry, leading to biased models. To address this, we introduce a method combining ResEmoteNet with diffusion-based data augmentation using Stable Diffusion \cite{stable-diff-2} to generate synthetic images from text prompts. This approach creates a more balanced dataset, enhancing the model's generalization across emotional expressions. Our experiments show that this method not only resolves class imbalance but also significantly improves FER accuracy, setting a new performance benchmark.

\section{ResEmoteNet Architecture}
\label{sec:methodology}
ResEmoteNet \cite{roy2024resemotenet} has an extensive architecture consisting of Squeeze and Excitation blocks and Residual blocks. These blocks help in minimizing losses while training and are capable of learning complex features resulting in a model that helps in accurate classification of emotions.

\begin{figure}
    \centering
    \hspace{-0.4cm}
\includegraphics[width=9cm,height=5.2cm]{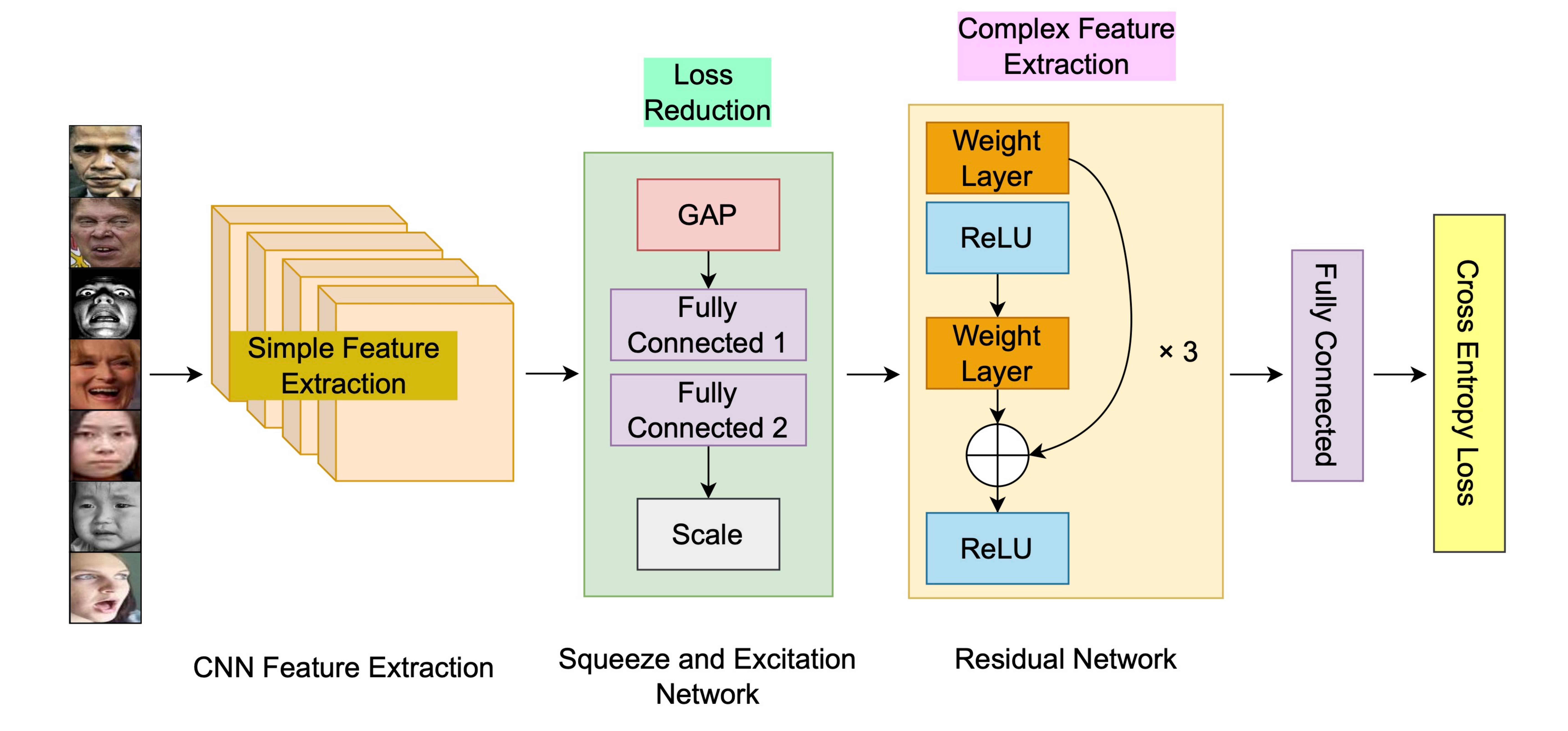}
    \label{fig: model-arch}
    \caption{Overview of ResEmoteNet for efficient facial emotion recognition.}
\end{figure}

\subsection{Loss Minimization}
The model architecture for ResEmoteNet is shown in Fig. \ref{fig: model-arch}, which consists of three parts, i.e., simple feature extraction from the CNN backbone, the Squeeze and Excitation Network (SENet) and the Residual Network for complex feature extraction. Given a sample $X \in \mathbb{R}^{H_0 \times W_0 \times 3}$ with RGB facial image of size $H_0 \times W_0$ ($H_0$ being the height of the facial image and $W_0$ being the width of the facial image), we utilize the CNN backbone to extract the simple features from the samples. The CNN backbone comprises of Convolution block accompanied by Batch Normalization to generate high-level feature maps of size $H \times W$ for each image. The spatial features $f_0 \in \mathbb{R}^{H \times W \times C}$ are then concatenated across channels to form a rich representation of the input image, which is further processed by the subsequent layers of the network. Subsequent to the extraction of high-level feature maps, a max-pooling layer is applied to further refine the feature representation. This layer reduces the spatial dimensions of the feature maps by retaining the most prominent features, thereby enhancing the model's efficiency and reducing the risk of overfitting.

The SE Block strengthens the network's ability to capture channel-wise features by applying a Global Average Pooling (GAP), condensing the spatial data which is following by a gating mechanism with Sigmoid activation to learn the attention weights. These weights modulate the spatial feature maps $f_0$ as described by:

\begin{equation}
    Y = w_s \cdot f_0
\end{equation}
where $w_s$ represents the attention weights derived from the SE Block. This results in a new feature map $f_1 \in \mathbb{R}^{H \times W \times C}$. It is noted that the temporal order of $f_1$ is in accordance with that of the input $X$.

\subsection{Residual Feature Extraction}
The Residual Network comprises three Residual Blocks, each with weight layers followed by ReLU activation and skip connections that iteratively learn residual functions. These functions model the differences between the block’s input and output, rather than unreferenced mappings. The skip connections bypass layers within a block, helping to train deeper architectures by mitigating vanishing gradients. This design accelerates model convergence and enhances generalization by preserving the original input signal throughout the network. The operation within a Residual block can be defined as:
\begin{equation}
    Y_{res} = H(f_1) + f_1
\end{equation}

where $H(f_1)$ is the output of the stacked layers, and $f_1$ is the input. Additionally, the network incorporates Adaptive Average Pooling (AAP) to ensure consistent output dimensions, irrespective of input size, across different datasets. The final stage of the network produces a probability distribution over facial emotion classes, computed as:
\begin{equation}
    P_{out} = f_{cls}(\tilde{Y}_{feats})
\end{equation}
where \(\tilde{Y}_{feats}\) is the output from the AAP operation, \(f_{cls}\) is the last layers of the architecture that gives the probability distribution over facial emotion classes and \(P_{out}\) is the classification result.

\section{Diffusion Based Data Augmentation}
\vspace{0.5em}
\begin{figure}
    \centering
    \includegraphics[width=\linewidth]{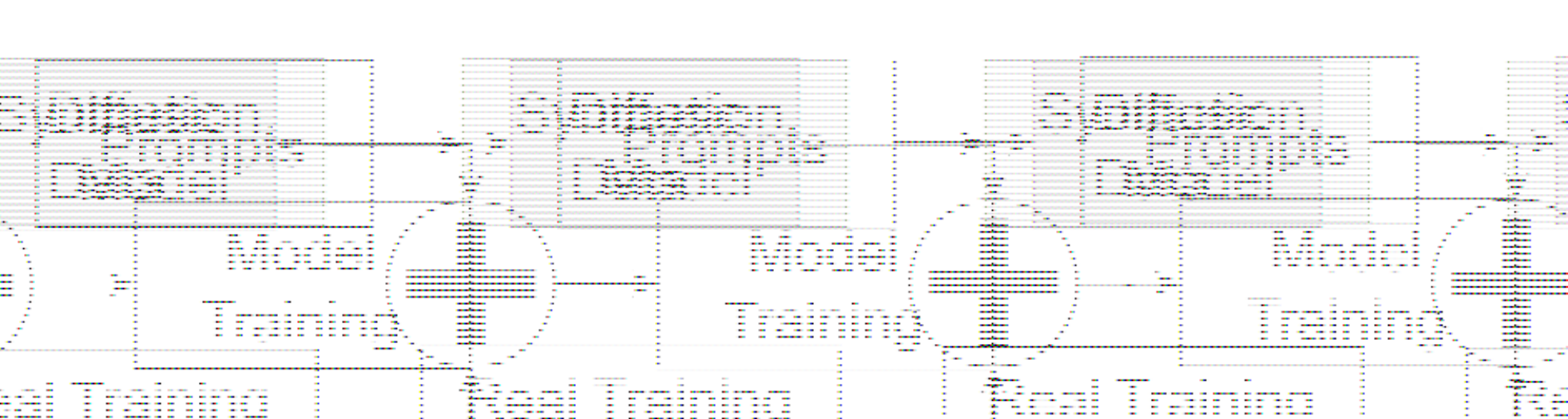}
    \caption{Overall pipeline for data augmentation.}
    \label{fig:pipeline}
\end{figure}

The class distribution in FER2013 and RAF-DB datasets varies significantly, leading to class imbalance. For instance, the Disgust class in FER2013 contains only 436 samples, while the Happy class has 7,215 samples. Similarly, in RAF-DB, the Fear class has just 281 samples compared to 4,772 samples in the Happy class. This imbalance causes the model to struggle with accurately classifying images from underrepresented classes, reducing the accuracy for those categories. To address this issue and balance the class distribution, we employ a generative diffusion model to generate additional facial images, as depicted in the pipeline shown in Fig. \ref{fig:pipeline}.

\vspace{0.7em}
\textbf{Diffusion Models: }Diffusion models are generative models that create high-quality images from textual descriptions by iteratively denoising a random image, gradually evolving noise into realistic data distributions. In this experiment, we used two such models: Stable-Diffusion-2 \cite{stable-diff-2} and Stable-Diffusion-3-Medium \cite{esser2024scaling}. Stable-Diffusion-2 incorporates Progressive Distillation \cite{salimans2022progressive} for faster sampling, trained on the LAION dataset \cite{schuhmann2021laion} for 550k steps at $256 \times 256$ resolution, and refined further to output $768 \times 768$ images. Stable-Diffusion-3-Medium, using the MM-DiT backbone built on Diffusion Transformers (DiTs) \cite{peebles2023scalable}, also starts with lower-resolution images, fine-tuned on ImageNet \cite{russakovsky2015imagenet} and CC12M \cite{changpinyo2021conceptual} datasets for 500k steps with a batch size of 4096. Both models generate images from text prompts by tokenizing and emphasizing key terms, with synthetic data generated using a variety of prompts, some of which are listed in Table \ref{tab:prompt-table}.

\renewcommand{\arraystretch}{1.5}
\begin{table}[!ht]
    \centering
    \caption{Sample prompts used as inputs for Diffusion models in generating synthetic data (not an exhaustive list).}
    \label{tab:prompt-table}
    \resizebox{\columnwidth}{!}{%
    \begin{tabular}{|p{0.7\linewidth}|p{0.3\linewidth}|}
         \hline
         \multicolumn{1}{|c|}{\textbf{Prompts}} & \multicolumn{1}{c|}{\textbf{Keywords}} \\
         \hline
         ``A man's face with surprise emotion" & \textbf{man, surprise} \\
         \hline
         ``A face of a woman in her 30's expressing disgust emotion" & \textbf{woman, 30's, disgust} \\
         \hline
         ``Happy expression on a kid, realistic photo" & \textbf{happy, kid, realistic} \\
         \hline
         ``An old man in his 80's expressing neutral emotions on his face" & \textbf{old man, 80's, neutral} \\
         \hline
    \end{tabular}%
    }
\end{table}

\begin{figure}[!ht]
    \centering
    \includegraphics[height=4.0cm]{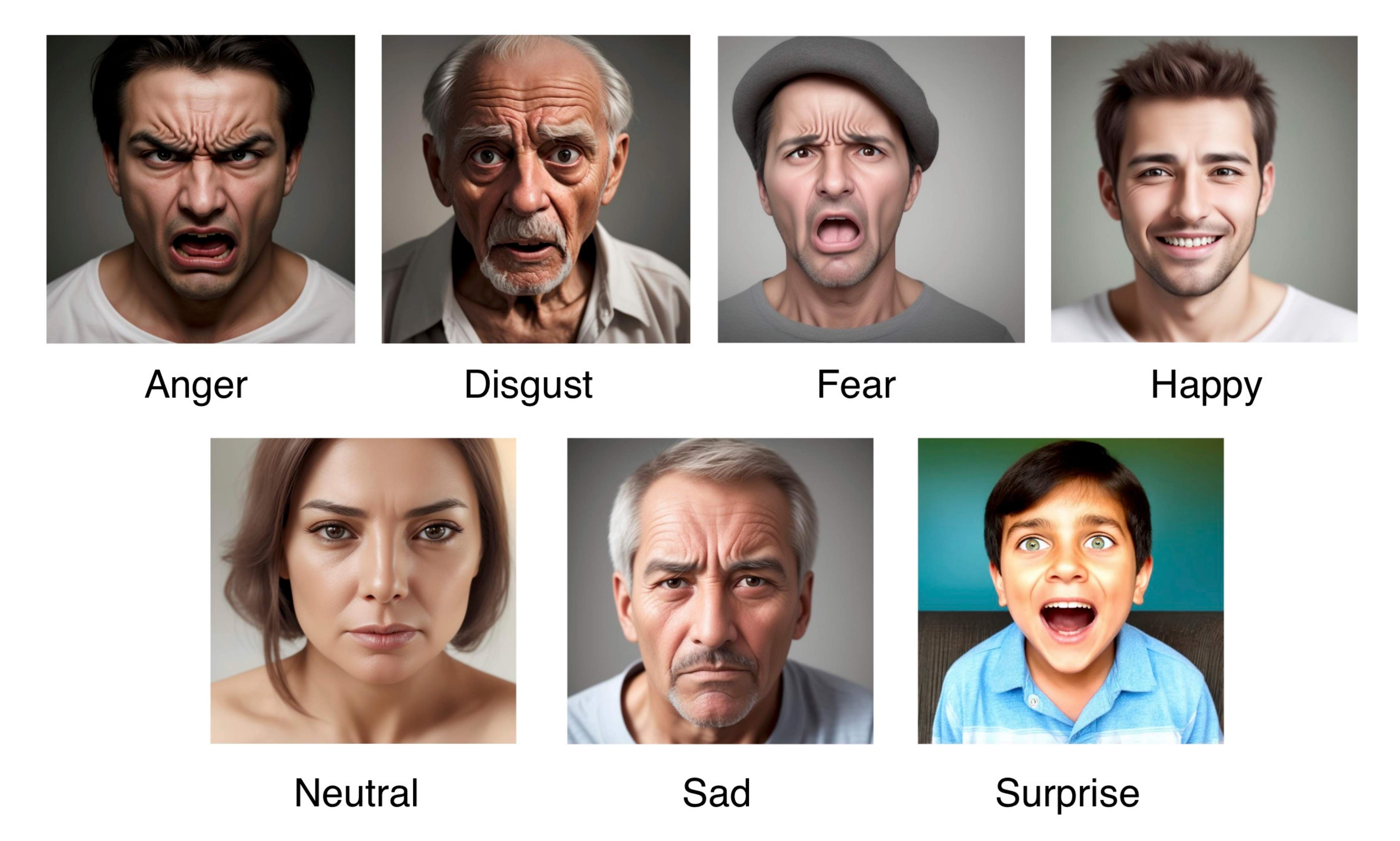}
    \label{fig:synthetic-samples}
    \caption{Examples of synthetic images representing each emotion class in the Facial Emotion Recognition (FER) task, generated using Diffusion Models based on the prompts listed in Table \ref{tab:prompt-table}.}
\end{figure}

Referring to the prompts listed in Table \ref{tab:prompt-table}, we generate synthetic facial images as shown in Fig. \ref{fig:synthetic-samples}. These synthetic images were resized to $64 \times 64$ to align with the input requirements of the model and subsequently integrated into the training set of the original datasets: FER2013 and RAF-DB. The validation and testing sets remained unchanged. The model was then trained on the augmented datasets, and its performance was closely monitored.

\vspace{0.3em}
\section{Dataset and Experimental Setup}
\label{sec:experiments}

\textbf{Datasets: }We evaluate our proposed methodology using two datasets: FER2013 \cite{goodfellow2015challenges} and RAF-DB \cite{li2017reliable}. The FER2013 dataset was created for a facial expression recognition challenge at the ICML Workshop in Representation Learning and contains 35,887 grayscale images of size ${48 \times 48}$. It is divided into training (28,709 images), public test (3,589 images), and private test sets (3,589 images), with annotations for seven basic emotions: Angry, Disgust, Fear, Happy, Sad, Surprise, and Neutral. FER2013 includes diverse images, such as occluded faces, low contrast, and subjects with eyeglasses. Similarly, the RAF-DB dataset, created in 2017, includes around 30,000 facial images annotated by 40 crowdsourced labelers. This research focuses on the basic single-label subset, comprising 12,271 training samples and 3,068 testing samples, with ${100 \times 100}$ images in three color channels, categorized into the same seven emotions as FER2013. Data distributions for both datasets are shown in Table \ref{tab:Augmented-Distribution-FER}.

\begin{table}[!ht]
\centering
\caption{Data distribution for training samples in FER2013 and RAF-DB datasets after augmentation. (Aug. refers to Augmentation)}
\label{tab:Augmented-Distribution-FER}
\scalebox{0.8}{ % Adjust table size to fit within the document
\begin{tabular}{|c|c|c|c|c|c|c|c|}
\hline
\multirow{2}{*}{\textbf{Classes}} & \multicolumn{2}{c|}{\textbf{Original}} & \multicolumn{2}{c|}{\textbf{Aug. 1}} & \textbf{Aug. 2} & \textbf{Aug. 3} & \textbf{Aug. 4} \\ \cline{2-8}
                                   & \textbf{FER2013} & \textbf{RAF-DB} & \textbf{FER2013} & \textbf{RAF-DB} & \textbf{Both} & \textbf{Both} & \textbf{Both} \\ \hline
\textbf{Anger}   & 3995          & 705            & 7215            & 4772            & 10000          & 12500          & 15000          \\ \hline
\textbf{Disgust} & 436           & 717            & 7215            & 4772            & 10000          & 12500          & 15000          \\ \hline
\textbf{Fear}    & 4097          & 281            & 7215            & 4772            & 10000          & 12500          & 15000          \\ \hline
\textbf{Happy}   & 7215          & 4772           & 7215            & 4772            & 10000          & 12500          & 15000          \\ \hline
\textbf{Neutral} & 4965          & 2524           & 7215            & 4772            & 10000          & 12500          & 15000          \\ \hline
\textbf{Sad}     & 4830          & 1982           & 7215            & 4772            & 10000          & 12500          & 15000          \\ \hline
\textbf{Surprise} & 3171         & 1290           & 7215            & 4772            & 10000          & 12500          & 15000          \\ \hline
\end{tabular}
}
\end{table}

\begin{figure*}
    \centering
    \label{fig:cm-label}
    \subfigure[\label{fig:fer_original}]
    {\includegraphics[width=0.238\textwidth]{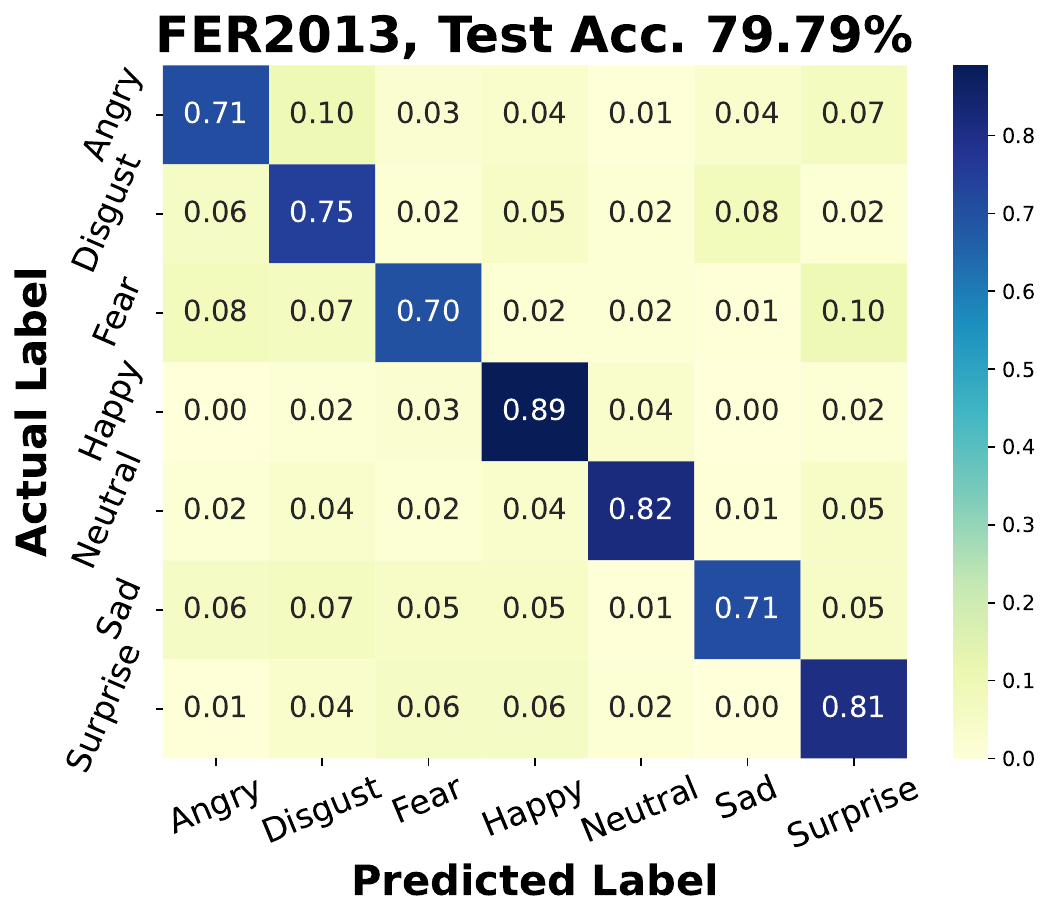}
    } \hfill
    \subfigure[\label{fig:fer_15k}]
    {\includegraphics[width=0.238\textwidth]{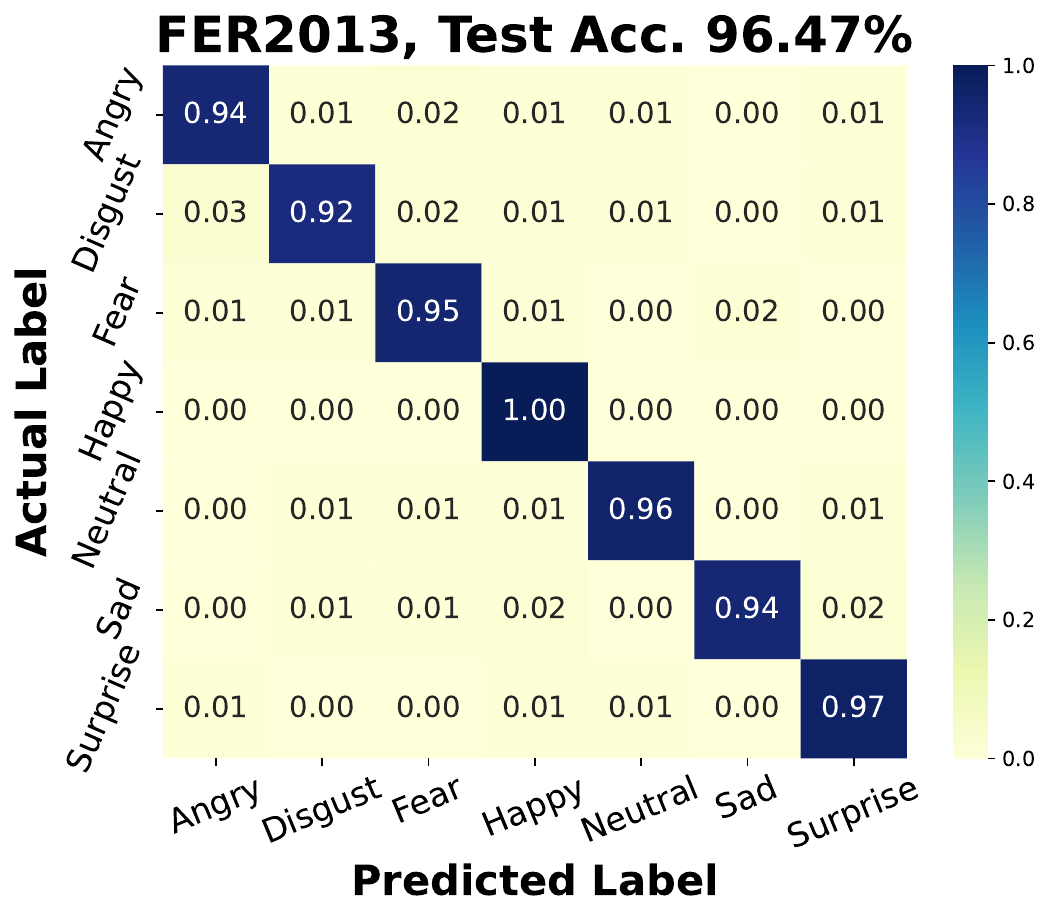}
    } \hfill
    \subfigure[\label{fig:raf_original}]
    {\includegraphics[width=0.238\textwidth]{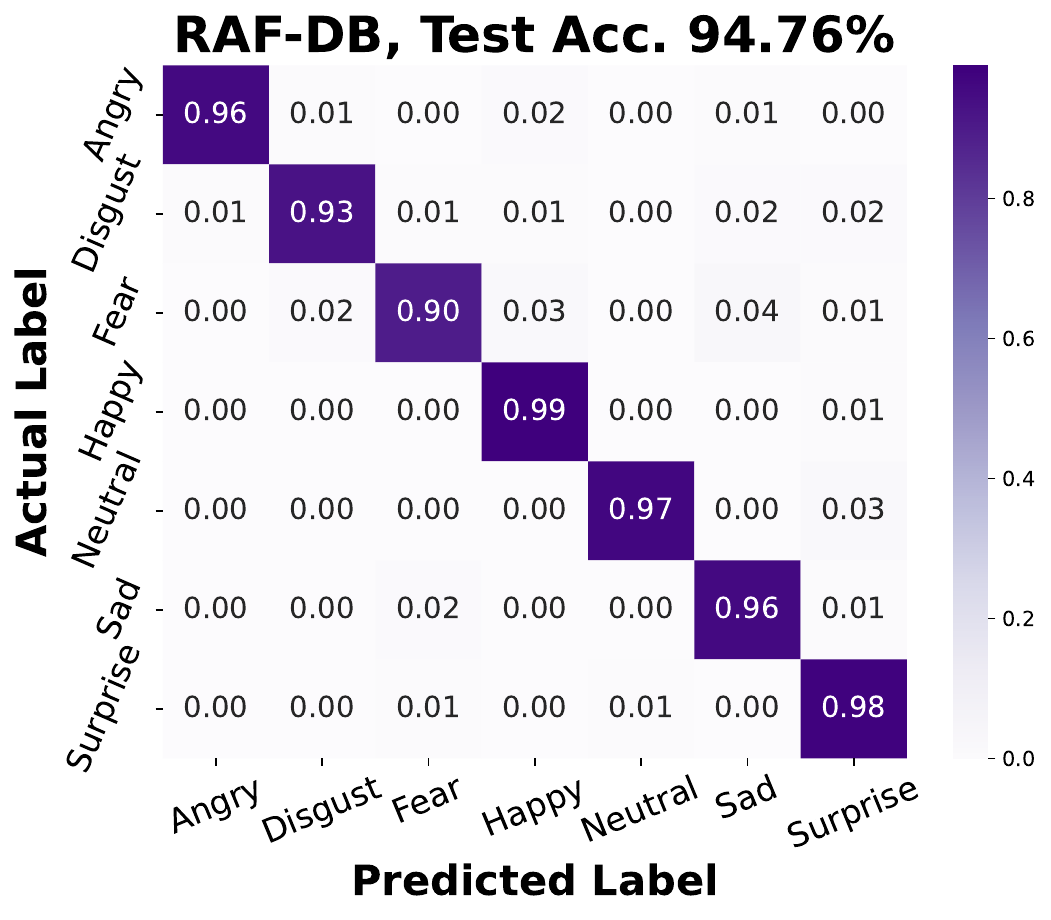}
    } \hfill
    \subfigure[\label{fig:raf_15k}]
    {\includegraphics[width=0.238\textwidth]{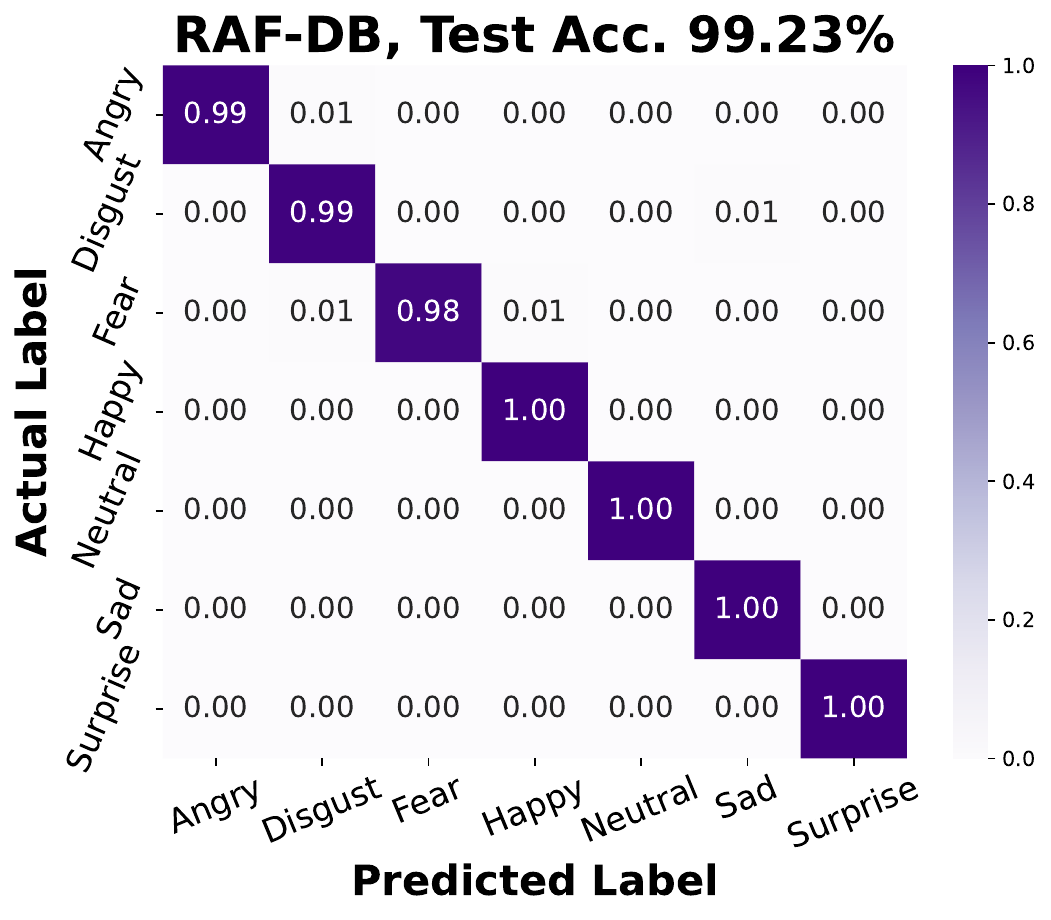}
    } \hfill
    \hfill
    \caption{Confusion matrices for the performance of the model on (a) Original FER2013, (b) FER2013 with Augmentation 4 (15000 samples in each class), (c) Original RAF-DB and (d) RAF-DB with Augmentation 4 (15000 samples in each class).}
\end{figure*}

\vspace{0.7em}
\textbf{Data Processing: }After the synthetic data were generated and added to the training sets of FER2013 and RAF-DB, each dataset was further divided into four augmentation subsets: Augmentation 1 (equal class distribution), Augmentation 2 (10,000 samples per class), Augmentation 3 (12,500 samples per class), and Augmentation 4 (15,000 samples per class).

\vspace{0.7em}
\textbf{Training Configuration: }We used the Stochastic Gradient Descent optimizer \cite{ruder2016overview} with an initial learning rate of $10^{-3}$, along with a learning rate scheduler that reduced by a factor of 0.1 upon reaching a plateau, and a batch size of 16. Cross-Entropy Loss was employed as the cost function \cite{zhang2018generalized}. ResEmoteNet was trained for up to 80 epochs with early stopping based on validation loss, with a patience of 5 epochs. The model converged before the maximum epochs on both datasets. Training with Augmentation 4 took 6.5 hours for FER2013 and 5 hours for RAF-DB, with inference times of 1.4 ms and 4.7 ms per image, respectively, using a batch size of 16. All training was conducted on a NVIDIA Tesla P100 GPU.

\begin{table}[!ht]
\centering
\caption{Class-wise test accuracy (\%) comparison of the model on FER2013 and RAF-DB datasets along different augmentations (Aug.).}
\label{tab:Accuracy-table}
\scalebox{0.85}{
\begin{tabular}{|c|c|c|c|c|c|c|}
\hline
\multirow{2}{*}{\textbf{Classes}} & \multirow{2}{*}{\textbf{Dataset}} & \multicolumn{5}{c|}{\textbf{Accuracy (\%)}} \\ \cline{3-7}
                                   &                                   & \textbf{Original} & \textbf{Aug. 1} & \textbf{Aug. 2} & \textbf{Aug. 3} & \textbf{Aug. 4} \\ \hline
\multirow{2}{*}{\textbf{Anger}}    & FER2013                           & 71                & 81              & 88              & 92              & 94              \\ \cline{2-7}
                                   & RAF-DB                            & 96                & 97              & 98              & 98              & 99              \\ \hline
\multirow{2}{*}{\textbf{Disgust}}  & FER2013                           & 75                & 80              & 88              & 91              & 92              \\ \cline{2-7}
                                   & RAF-DB                            & 93                & 96              & 98              & 99              & 99              \\ \hline
\multirow{2}{*}{\textbf{Fear}}     & FER2013                           & 70                & 82              & 90              & 93              & 95              \\ \cline{2-7}
                                   & RAF-DB                            & 90                & 94              & 97              & 98              & 98              \\ \hline
\multirow{2}{*}{\textbf{Happy}}    & FER2013                           & 89                & 94              & 97              & 99              & 100             \\ \cline{2-7}
                                   & RAF-DB                            & 99                & 100             & 100             & 100             & 100             \\ \hline
\multirow{2}{*}{\textbf{Neutral}}  & FER2013                           & 82                & 87              & 91              & 95              & 96              \\ \cline{2-7}
                                   & RAF-DB                            & 97                & 98              & 100             & 100             & 100             \\ \hline
\multirow{2}{*}{\textbf{Sad}}      & FER2013                           & 71                & 81              & 88              & 93              & 94              \\ \cline{2-7}
                                   & RAF-DB                            & 96                & 97              & 99              & 99              & 100             \\ \hline
\multirow{2}{*}{\textbf{Surprise}} & FER2013                           & 81                & 88              & 93              & 96              & 97              \\ \cline{2-7}
                                   & RAF-DB                            & 98                & 99              & 100             & 100             & 100             \\ \hline \hline
\multirow{2}{*}{\textbf{Overall Accuracy}} & FER2013                   & 79.79             & 84.81           & 91.48           & 94.69           & \textbf{96.47}  \\ \cline{2-7}
                                   & RAF-DB                            & 94.76             & 96.17           & 98.58           & 98.91           & \textbf{99.23}  \\ \hline
\end{tabular}
}
\end{table}

\section{Results and Discussion}
\label{sec:results}

Table \ref{tab:Accuracy-table} demonstrates the substantial improvements in model accuracy on both the FER2013 and RAF-DB datasets following various augmentations. On FER2013, the accuracy rose significantly from \textbf{79.79\%} to \textbf{96.47\%}, with notable gains in the Fear class (25\% increase) and the Happy class (11\% increase), leading to a perfect 100\% accuracy in the latter. Similarly, the Sad and Neutral classes saw substantial improvements, further enhancing overall performance. On RAF-DB, the model’s accuracy increased from an initial \textbf{94.76\%} to \textbf{99.23\%} after the final augmentation, with the Disgust class improving from 93\% to 99\%, and perfect accuracy achieved in the Happy, Neutral, Sad, and Surprise classes. The Fear class also saw an 8\% improvement, reaching 98\%, ensuring that the model consistently performed well across all categories.

\subsection{Comparison with previous study}

Data augmentation, particularly through synthetic data generated by diffusion models, proved crucial in addressing class imbalances and improving facial emotion recognition. By increasing the diversity and quantity of training data, this strategy significantly enhanced ResEmoteNet’s performance, especially in challenging classes like Fear and Disgust. As shown in Table \ref{tab:result-compare}, data augmentation enabled our model to surpass state-of-the-art methods, demonstrating the importance of such techniques in boosting model robustness, accuracy, and generalizability.
\begin{comment}
\begin{table}[!ht]
    \centering
    \caption{Test accuracy (\%) comparison of ResEmoteNet on FER2013 and RAF-DB datasets along different augmentations. The best results are in bold.}
    \label{tab:Accuracy-table}
    \vspace{1em}
    \begin{tabular}{|c|c|c|}
    \hline
    \textbf{Distribution} & \multicolumn{2}{c|}{\textbf{Accuracy (\%)}} \\
    \cline{2-3} & \textbf{FER2013} & \textbf{RAF-DB} \\ 
    \hline
    Original & 79.79 & 94.76 \\ \cline{2-3} 
    \hline
    Augmentation 1 & 84.81 & 96.17 \\ \cline{2-3}
    \hline
    Augmentation 2 & 91.48 & 98.58 \\ \cline{2-3}
    \hline
    Augmentation 3 & 94.69 & 98.91 \\ \cline{2-3}
    \hline
    Augmentation 4 & \textbf{96.17} & \textbf{99.23} \\ \cline{2-3}
    \hline
    \end{tabular}
\end{table}
\end{comment}

\renewcommand{\arraystretch}{1.5}
\begin{table}[!ht]
    \centering
    \caption{Test Accuracy (\%) comparison with existing state-of-the-art methods across: FER2013 and RAF-DB dataset.}
    \label{tab:result-compare}
    \vspace{0.3em}
    \small
    \scalebox{0.85}{
    \setlength{\tabcolsep}{10pt}
    \begin{tabular}{|c|c|c|}
        \hline
           & \multicolumn{2}{|c|}{\bf {Accuracy in \%}} \\ \cline{2-3}
            \textbf{Method} & \textbf{FER2013} & \textbf{RAF-DB}\\
        \hline
        LANMSFF \cite{ezati2024lightweight} & 70.44 & - \\
        \hline
         Local Learning Deep+BOW \cite{georgescu2019local} & 75.42 & - \\
         \hline
         Multi-Branch ViT \cite{zhu2024cross} & 77.85 & 64.28 \\
         \hline
         EmoNeXt \cite{el2023emonext} & 76.12 & - \\
          \hline
         Ensemble ResMaskingNet \cite{residualmasking} & 76.82 & - \\
          \hline
         MixCut \cite{yu2024mixcut} & - & 86.84 \\
         \hline
         DDAMFN++ \cite{zhang2023dual} & - & 92.34 \\
          \hline
         S2D \cite{s2d} & - & 92.57   \\
         \hline
         FMAE \cite{ning2024representation} & - & 93.09 \\
         \hline
         \hline
         \textbf{ResEmoteNet (Without Augmentation)} & \textbf{79.79} & \textbf{94.76} \\
         \hline
         \hline
         \textbf{ResEmoteNet (With augmentation)} & \textbf{96.47} & \textbf{99.23} \\
         \hline
    \end{tabular}}
\end{table}

\vspace{0.5cm}
\section{Conclusion}
\label{sec:conclusion}
\vspace{0.3cm}

This study highlights the importance of data augmentation in enhancing the performance of neural networks for facial emotion recognition. By employing advanced generative models like Stable Diffusion 2 and 3, we showed that synthetic data is instrumental in mitigating class imbalances and significantly boosting the overall robustness of neural networks. The significant improvements observed across both FER2013 and RAF-DB benchmarks validate the efficacy of our approach. Notably, ResEmoteNet’s accuracy on the FER2013 dataset improved from 79.79\% to 96.47\% with augmentation, and on the RAF-DB dataset, the model’s accuracy increased from 94.76\% to 99.23\%, marking substantial gains over the previous state-of-the-art.

\newpage

\bibliographystyle{ieeetr}
\bibliography{main.bib}

\end{document}